\documentclass[10pt,twocolumn,letterpaper]{article}
\usepackage[pagenumbers]{cvpr}
\ifdefined\pdfinfoomitdate\pdfinfoomitdate=1\relax\fi
\ifdefined\pdfsuppressptexinfo\pdfsuppressptexinfo=15\relax\fi
\usepackage{mathtools}
\usepackage{microtype}
\usepackage{booktabs}
\usepackage{graphicx}
\usepackage{placeins}
\usepackage{algorithm}
\usepackage{algpseudocode}
\newcommand{\method}{DiE-VL}
\newcommand{\E}{\mathbb{E}}

\newcommand{\Sim}{\operatorname{Sim}}
\newcommand{\clip}{\operatorname{clip}}

\definecolor{cvprblue}{rgb}{0.21,0.49,0.74}
\usepackage[breaklinks,colorlinks,allcolors=cvprblue]{hyperref}
\hypersetup{
  pdftitle={DiE-VL: Robust and Transferable Defense against Image Editing via VLM Perturbation},
  pdfauthor={Qichen Zhao, Shiqi Zheng, Wenjun Ji, Pengfei Wu, Shengfang Zhai, Qingni Shen}
}
\title{DiE-VL: Robust and Transferable Defense against Image Editing\\via VLM Perturbation}
\author{%
  Qichen Zhao$^{1*}$ \qquad Shiqi Zheng$^{1*}$ \qquad Wenjun Ji$^{1}$ \\
  Pengfei Wu$^{1}$ \qquad Shengfang Zhai$^{2\dagger}$ \qquad Qingni Shen$^{1\dagger}$ \\[3pt]
  {\normalsize $^{1}$Peking University \qquad $^{2}$National University of Singapore} \\
  {\small \href{mailto:z24@stu.pku.edu.cn}{z24@stu.pku.edu.cn}}%
}
\begin{document}
\maketitle
{\renewcommand{\thefootnote}{\fnsymbol{footnote}}%
\footnotetext[1]{Equal contribution.}%
\footnotetext[2]{Corresponding authors.}%
}
\begin{figure*}[!t]
  \centering
  \includegraphics[width=\textwidth]{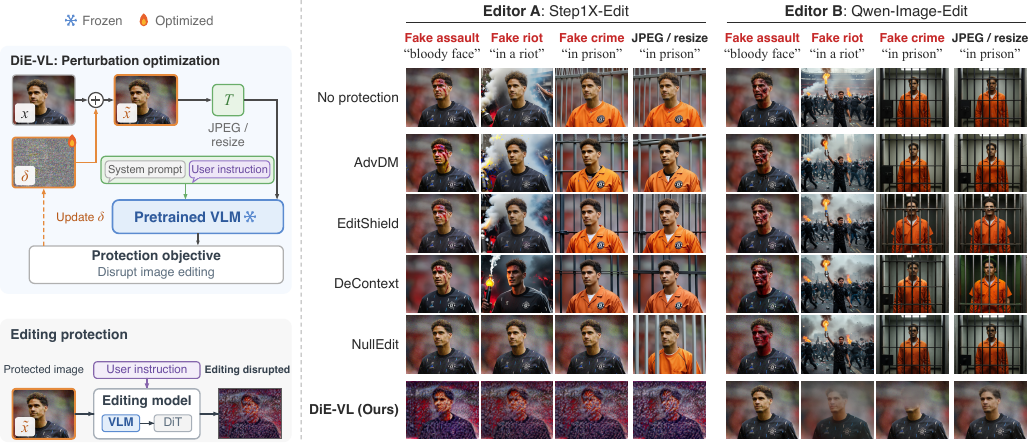}
  \caption{\textbf{Protection workflow and qualitative results of \method{}.} The left panel shows \method{} optimizing a perturbation on a frozen pretrained VLM and reusing the protected image across editing models. The right panel compares protection methods on Step1X-Edit and Qwen-Image-Edit, including results after JPEG compression or resizing.}
  \label{fig:teaser}
\end{figure*}
\begin{abstract}
Image editing methods based on vision--language models (VLMs) combine reference images and user instructions to guide generation, making image modification more accessible while exposing public images to unauthorized manipulation.
Existing defenses primarily target image encoding and denoising in diffusion models, whereas VLM-oriented protection fails to reliably suppress edits when a single protected image is transferred across editing models or subjected to compression and resizing.
We introduce \method{}, which optimizes image perturbations using only a pretrained VLM, without accessing downstream generative model parameters, so that one protected image per original can be directly reused across editing models.
Using the original-image representations as references, \method{} jointly disrupts the VLM's encoding of images and editing instructions, strengthens this disruption through intermediate-layer supervision, and adaptively selects system prompt and user editing instruction pairs for optimization based on final-layer instruction representation similarity.
To maintain effectiveness across model pipelines and image sharing, we jointly optimize a single perturbation over diverse visual preprocessing grids, JPEG compression, and resizing.
In experiments with 100 images and 200 tasks with unseen instructions, the same set of protected images is directly used on Step1X-Edit and Qwen-Image-Edit, reducing instruction adherence by at least 71.6\% relative to the strongest baseline on each model and yielding smaller perceptual differences between the protected and original images than all four baselines.
After JPEG compression or resizing, the same protected images still yield the lowest instruction adherence across all tested conditions.
\end{abstract}

\section{Introduction}
\label{sec:introduction}
Image editing models allow users to specify desired changes in natural language and modify people's appearance, object attributes, and scenes in reference images.
Editing methods using vision--language models (VLMs)~\cite{step1x,qwenimage} further combine reference-image content with user instructions to identify editing targets and provide semantic conditions for image generation.
This ease of editing also exposes public images to unauthorized manipulation~\cite{veto}, as third parties can combine acquired images with their own instructions to alter a person's appearance or generate misleading content.
Protection methods applied before publication~\cite{photoguard,dctshield} embed visually subtle perturbations into the original image to proactively interfere with subsequent unauthorized editing.
After publication, owners have little control over downstream editing models and instructions~\cite{tdae}, and compression and resizing during dissemination~\cite{dctshield} can further alter the images received by editing models.
The same protected image constructed before publication must therefore suppress editing across different models and instructions and remain effective after dissemination processing.

Existing image editing protection methods construct perturbations using image encodings~\cite{editshield}, denoising processes~\cite{photoguard,sdsprotect}, or reference attention in diffusion transformers (DiTs)~\cite{decontext} within specific models.
Research on transferable protection~\cite{tdae,unidef} improves perturbation effectiveness across generative models, while compression-robust protection~\cite{dctshield} strengthens protection after JPEG processing, but these methods primarily optimize against image encoding and generation in diffusion models.
VLM-based editing models~\cite{step1x,qwenimage} jointly encode reference images and user editing instructions under system prompts to provide semantic conditions for generation, so protected images must also accommodate differences in system prompts and visual preprocessing across models.
VLM-based protection~\cite{nulledit} suppresses modifications by disrupting the joint encoding of images and editing instructions.
However, existing defenses degrade markedly when a single protected image is transferred to editors with different system prompts and visual preprocessing pipelines, or subjected to compression and resizing during sharing.
We therefore investigate whether perturbations optimized solely on a pretrained VLM can enable the same published image to suppress modifications by different downstream editing models after common image processing during dissemination.

Achieving this goal requires addressing three challenges.
First, open-source editing models~\cite{step1x,qwenimage,uniworld,omnigen2} use a pretrained VLM to jointly encode reference images and user instructions, but differ in both visual preprocessing and downstream conditioning interfaces.
Specifically, visual preprocessing varies the input resolution and the number and layout of visual tokens~\cite{qwen2vl}, while each editor applies its own rescaling and projection to map VLM hidden states into generation conditions~\cite{step1x,qwenimage}.
Consequently, preprocessing alters how a pixel perturbation shifts VLM representations, and editor-specific interfaces govern how those shifts propagate to the diffusion backbone.
Perturbations optimized on the VLM alone must therefore induce representation shifts that remain disruptive across distinct conditioning interfaces.
Second, editing models may supply different system prompts to the VLM.
For the same image and user editing instruction, a change in the system prompt can alter the VLM's hidden states and affect the perturbation's effectiveness.
The same protected image therefore needs to suppress edits across combinations of system prompts and user editing instructions.
Finally, JPEG quantization~\cite{dctshield} may remove some small perturbation components, while interpolation during resizing changes perturbations' spatial distribution.
The visual quality required of the published image also limits perturbation magnitude, so protection must suppress edits after compression or resizing within a limited pixel budget.

We propose \method{}, a defense against image editing that optimizes image perturbations solely on a pretrained vision--language model without accessing downstream generative model parameters (Figure~\ref{fig:teaser}).
\method{} shifts both image and instruction representations away from the corresponding representations of the original image and uses intermediate-layer supervision to strengthen the disruption of generation conditions.
We optimize the same perturbation across multiple system prompts, user editing instructions, and visual preprocessing settings, while incorporating JPEG compression and resizing to accommodate changes in editing inputs and dissemination processing.
During optimization, we track final-layer instruction similarity between the protected and original images to prioritize system-prompt and instruction pairs that exhibit smaller directional shifts.
On 100 images and 200 tasks with unseen instructions, the same set of protected images is directly reused on Step1X-Edit and Qwen-Image-Edit, reducing instruction adherence by 80.7\% and 71.6\% relative to the strongest baseline on each model, respectively, while yielding smaller perceptual differences between the protected and original images than all baselines.
After JPEG compression or resizing, the same protected images still yield the lowest instruction adherence across all tested conditions.
Our main contributions are:
\begin{itemize}
    \item We introduce \method{}, which uses only a pretrained VLM to construct protected images reusable across editing models, with protection retained after compression or resizing.
    \item We design a multilayer representation loss that moves both image and instruction representations of the protected image away from their counterparts for the original image at intermediate and final VLM layers. We further introduce similarity-based instruction selection, which adaptively reweights system-prompt and instruction pairs according to their final-layer instruction similarity.
    \item In experiments with 100 images, 200 tasks with unseen instructions, and two editing models, \method{} reduces instruction adherence by at least 71.6\% relative to the strongest baseline on each model and yields smaller perceptual differences between the protected and original images than all four baselines.
\end{itemize}

\section{Related Work}
\label{sec:related}

\subsection{Conditional Image Editing}
Diffusion-based controllable generation methods~\cite{ldm} use spatial conditions~\cite{controlnet} or reference-image features~\cite{ipadapter} to constrain the structure and appearance of generated images.
Text-guided image editing methods adapt to reference images by optimizing text embeddings and model parameters~\cite{imagic}, or constrain editing content and regions through attention control and local spatial constraints~\cite{p2p,blended}.
InstructPix2Pix~\cite{instructpix2pix} is trained with paired images and editing instructions to directly perform natural-language edits at inference time.
MGIE~\cite{mgie} and SmartEdit~\cite{smartedit} use multimodal large language models to interpret editing requests in the context of image content, and use the resulting semantic representations to guide diffusion generation.
Among editing models with Transformer-based generative backbones~\cite{dit,mmdit}, Step1X-Edit~\cite{step1x}, Qwen-Image-Edit~\cite{qwenimage}, UniWorld~\cite{uniworld}, and OmniGen2~\cite{omnigen2} use pretrained VLMs to encode reference images and user instructions, providing semantic conditions for downstream generation.
These methods improve models' ability to understand and execute editing instructions, but also increase the risk of unauthorized modifications to publicly available images.

\subsection{Proactive Image Protection}
Proactive image protection adds perturbations before publication to disrupt subsequent unauthorized generation and editing.
Protection against content and style imitation~\cite{advdm,sdsprotect} uses perturbations to impede diffusion models' use of reference-image features, while protection against subject personalization~\cite{metacloak,simac} adds perturbations to training images to limit learning of a specific subject's appearance.
For inference-time editing, existing methods construct perturbations targeting image encoding~\cite{photoguard}, latent representations~\cite{editshield}, or early denoising stages~\cite{diffusionguard}, and also induce deviations from the original image by disrupting identity preservation~\cite{facelock} or reference information propagation in DiTs~\cite{decontext}.
To improve protection across models, TDAE~\cite{tdae} combines image-gradient regularization with text-embedding optimization, while UniDef~\cite{unidef} uses finite differences to estimate gradients of the complete denoising process, reducing reliance on model-specific gradients.
To retain protection after compression, DCT-Shield~\cite{dctshield} optimizes image perturbations in the DCT domain.
For VLM-conditioned editing models, NullEdit~\cite{nulledit} uses VLM representations under edit and no-edit instructions as references to suppress edits while preserving the original image's content in the generated output.
Nevertheless, existing defenses remain ineffective when a single protected image must withstand both cross-editor configuration shifts and common dissemination transforms.

\subsection{Adversarial Perturbations against VLMs}
Adversarial attacks against VLMs jointly perturb image and text inputs~\cite{coattack,vlattack}, or target pretrained vision encoders~\cite{vlmrobust,veattack}, to disrupt tasks such as image--text retrieval, visual question answering, and image captioning.
Visual perturbations can also induce targeted responses~\cite{imagehijacks} or bypass models' safety alignment~\cite{visualjailbreak,jailbreakpieces}.
To improve cross-model transferability, existing work uses image--text semantic correlations~\cite{coa} and feature matching in local regions~\cite{mattack} to guide perturbation optimization, or trains perturbation generators through self-supervised learning~\cite{anyattack}.
To address prompt variations, InstructTA~\cite{instructta} uses instruction paraphrasing to increase the diversity of language inputs.
These studies construct perturbations targeting VLM predictions and textual responses, and their optimization objectives do not include suppressing downstream image editing.

\section{Preliminaries}
\label{sec:preliminaries}
\subsection{VLM-conditioned Image Editing}
\label{sec:vlm-editing}
VLM-based image editing models use a pretrained VLM to encode the reference image and editing instruction, and use its outputs to condition a diffusion model.
The reference image is also encoded into a latent representation by a variational autoencoder (VAE), providing the original image's structure and appearance for generation.
Given a reference image $x$ and editing instruction $u$, the VLM first extracts image features through its visual encoder, then feeds them to the language model together with the text embeddings of the system prompt $s$ and editing instruction.
The system prompt provides the task context, the editing instruction specifies the requested changes, and visual preprocessing determines the input image resolution.
The language model uses attention to model relationships between the image and text, producing hidden states that guide editing.
Let $F_\theta$ denote the encoding function of the pretrained VLM.
Editing model $e$ uses a mapping $\Phi_e$ to convert these hidden states into generation conditions $c_e$,
\begin{equation}
c_e=\Phi_e\!\left(F_\theta(x;s,u)\right).
\label{eq:vlm-final}
\end{equation}
This mapping adapts the VLM outputs to the input format of the generative model.
For example, Step1X-Edit~\cite{step1x} uses a connector to construct conditions for its diffusion model, while Qwen-Image-Edit~\cite{qwenimage} maps VLM representations to the text stream of its multimodal diffusion transformer (MMDiT).

The reference latent representation produced by the VAE encoder $\mathcal E_e$ is concatenated with the latent representation to be generated, which is initialized from noise $\xi$.
Conditioned on $c_e$, the diffusion model iteratively updates the generated part, which is then decoded by the VAE into the edited image $y$.
Denoting iterative denoising and decoding by $G_e$, we write
\begin{equation}
y=G_e\!\left(\xi;c_e,\mathcal E_e(x)\right).
\label{eq:editing-process}
\end{equation}

\subsection{Threat Model}
\label{sec:threat}
\paragraph{Scenario.}
We consider image editing models that use the same pretrained VLM to encode images and editing instructions.
An image owner adds perturbations before publication~\cite{photoguard} and releases one protected image for each original.
Downstream users can choose editing models and instructions~\cite{tdae} to perform unauthorized modifications, while the chosen model determines the system prompt and visual preprocessing.
JPEG compression and resizing during dissemination~\cite{dctshield} may further alter the image received by the model.
All subsequent edits use the corresponding protected image or a processed version of it.

\paragraph{Protector Capabilities and Objectives.}
The protector holds the original image and can access the parameters and preprocessing pipeline of a pretrained VLM to extract hidden states and compute input gradients.
The protector modifies the image within a given pixel budget while keeping the VLM parameters fixed, without accessing downstream generative model parameters.
The goal is to disrupt unauthorized edits while keeping the published image visually close to the original.
(1) Effectiveness requires preventing the editing model from fulfilling the user's instruction, either by blocking the requested modification or by corrupting the generated content.
(2) Transferability requires the same protected image to be reusable across editing models and to generalize to system prompts and editing instructions unseen during optimization.
(3) Robustness requires the protected image to continue suppressing unauthorized edits after JPEG compression or resizing.

\section{Methodology}
\label{sec:method}

\begin{figure*}[!t]
\centering
\includegraphics[width=\textwidth]{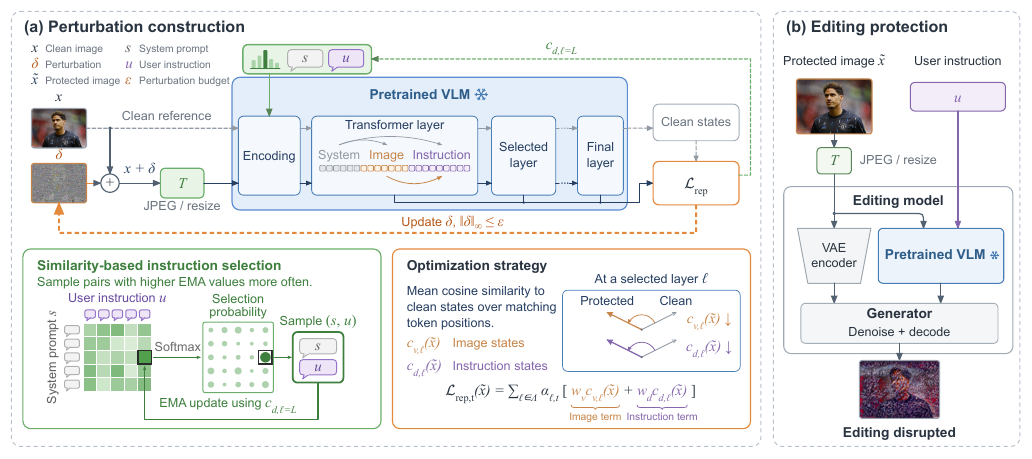}
\caption{\textbf{Overview of \method{}.} (a) A frozen pretrained VLM guides perturbation optimization. System--instruction pairs with higher exponential moving average (EMA) similarity scores are sampled more frequently. For $T(\tilde x)$ and clean $x$ under the same prompt pair and visual-token grid, $c_{v,\ell}$ and $c_{d,\ell}$ average cosine similarities over corresponding image and text positions. We minimize their weighted sum across layers $\Lambda$, with image/instruction weights $w_v,w_d$ and layer weights $\alpha_{\ell,t}$. In the figure, $\tilde x$ abbreviates $T(\tilde x)$, and $c_{d,\ell=L}$ denotes the final-layer instruction similarity. (b) The protected image is reused across editing models after compression or resizing.}
\label{fig:method-overview}
\end{figure*}

Figure~\ref{fig:method-overview} illustrates the perturbation optimization and cross-editor reuse pipeline of \method{}.
Section~\ref{sec:method-formulation} formalizes the protection objective and multilayer representation loss, Section~\ref{sec:conditioning} describes optimization across conditioning inputs and image transformations, and Section~\ref{sec:optimization} introduces similarity-based instruction selection.
Algorithm~\ref{alg:protection} summarizes the complete procedure.

\subsection{Problem Formulation}
\label{sec:method-formulation}
\label{sec:representations}

\paragraph{Protection Process.}
Given an $H\times W$ original image $x\in[0,1]^{H\times W\times3}$, \method{} optimizes a pixel perturbation $\delta$ within budget $\varepsilon$ to construct a protected image
\begin{equation}
 \tilde x=\clip_{[0,1]}(x+\delta),
 \qquad \|\delta\|_\infty\leq\varepsilon.
 \label{eq:protected}
\end{equation}
Optimization updates only $\delta$, keeping the parameters of the pretrained VLM $F_\theta$ fixed and requiring no access to downstream generative model parameters.
We construct one protected image per original image and use it across editing models without further optimization.
The protection goal is to accommodate changes in system prompts, editing instructions, and visual-token grids while continuing to suppress unauthorized modifications after compression and resizing.

Let $a=(s,u)$ denote the system prompt and user editing instruction pair, and let $r$ denote the visual-token grid.
Let $L$ denote the final layer of the language model, and $F_\theta^{(\ell)}$ the VLM encoding function through layer $\ell$ of the language model.
The corresponding hidden-state sequence is
\begin{equation}
 h^{(\ell)}(x;a,r)=F_\theta^{(\ell)}(x;a,r).
 \label{eq:vlm-layer}
\end{equation}
Within this sequence, the hidden states at image-token positions form the image representations $h_v^{(\ell)}(x;a,r)$.
The hidden states at all non-padding positions after the image-end marker form the instruction representations $h_d^{(\ell)}(x;a,r)$, covering the user editing instruction and subsequent dialogue-template tokens.

\paragraph{Optimization Problem.}
Using the original image's representations under the same $a,r$ as references, we reduce the representation similarity between the protected and original images to disrupt the VLM's encoding of the image and editing instruction.
Causal attention allows image information to affect the hidden states at subsequent text positions, so pixel perturbations can alter both groups of representations.
We use $k\in\{v,d\}$ to index the image and instruction representations.
Omitting the shared $a,r$, the layer-$\ell$ loss is
\begin{equation}
\begin{aligned}
 c_{k,\ell}(\tilde x)
 &=\Sim\!\left(h_k^{(\ell)}(\tilde x),h_k^{(\ell)}(x)\right),\\
 \mathcal L_\ell(\tilde x)
 &=w_vc_{v,\ell}(\tilde x)+w_dc_{d,\ell}(\tilde x).
\end{aligned}
\label{eq:condition_loss}
\end{equation}
Here $\Sim$ computes cosine similarities between hidden vectors at corresponding positions and averages them within each group.
The weights $w_v,w_d$ correspond to the image and instruction losses, respectively, with $w_v,w_d\geq0$ and $w_v+w_d=1$.
Final-layer supervision targets the encoding outputs used to construct generation conditions, while intermediate-layer supervision further reduces the representation similarity between the protected and original images during encoding.
With $\Lambda$ denoting the set of supervised layers (including $L$), the multilayer loss at iteration $t$ is
\begin{equation}
 \mathcal L_{\mathrm{rep},t}(\tilde x;a,r)
 =\sum_{\ell\in\Lambda}\alpha_{\ell,t}\mathcal L_\ell(\tilde x;a,r).
 \label{eq:layer_loss}
\end{equation}
The weight $\alpha_{\ell,t}$ of the layer-$\ell$ loss at iteration $t$ satisfies $\alpha_{\ell,t}\geq0$ and $\sum_{\ell\in\Lambda}\alpha_{\ell,t}=1$.

We compute this loss under different system prompts, editing instructions, visual-token grids, and image transformations.
Let $\mathcal S$ and $\mathcal U$ denote the sets of system prompts and editing instructions, respectively, forming the training set $\mathcal A=\mathcal S\times\mathcal U$, and let $\mathcal R$ denote the candidate set of visual-token grids.
The image transformation $T$ is either the identity, JPEG compression, or resizing; Appendix~\ref{sec:optimization-settings} details the transformations and surrogate gradients.
For $a\in\mathcal A$, $r\in\mathcal R$, and the chosen transformation $T$, the loss for the transformed protected image is
\begin{equation}
 \ell_{a,t}(\delta;r,T)
 =\mathcal L_{\mathrm{rep},t}\!\left(T(\tilde x);a,r\right).
 \label{eq:transformed_loss}
\end{equation}
Following Athalye et al.~\cite{eot}, we optimize the same perturbation using the expected representation loss over different $a,r,T$.
Let $\mu_t$ denote their joint probability distribution at iteration $t$, defined in Appendix~\ref{sec:effective-distribution}.
Within the perturbation budget, \method{} minimizes
\begin{equation}
\begin{aligned}
 \mathcal J_t(\delta)
 &=\E_{(a,r,T)\sim\mu_t}
   \left[\ell_{a,t}(\delta;r,T)\right]\\
 &\quad+\lambda_t\Omega(\delta),
 \qquad \|\delta\|_\infty\leq\varepsilon.
\end{aligned}
\label{eq:eot}
\end{equation}
The pixel regularizer $\Omega(\delta)=\|\tilde x-x\|_2^2/(3HW\varepsilon^2)$ controls pixel changes in the protected image relative to the original, with weight $\lambda_t$.

\subsection{\texorpdfstring{\raggedright Optimization for Transferability and Robustness}{Optimization for Transferability and Robustness}}
\label{sec:conditioning}
\label{sec:transformations}
Varying the system prompt, editing instruction, or visual-token grid changes both the VLM's encoding of a reference image and how a perturbation shifts its hidden states.
To generalize across editing models, each iteration samples a prompt--instruction pair $a=(s,u)\in\mathcal A$ via similarity-based selection (Section~\ref{sec:optimization}) and a visual-token grid $r\in\mathcal R$ to update the shared perturbation $\delta$.
To withstand dissemination processing, we progressively incorporate resizing and JPEG compression into the stage schedule (Table~\ref{tab:schedule}): at each step, we sample a transformation $T$ according to the current stage distribution and evaluate Eq.~\eqref{eq:transformed_loss} by comparing the hidden states of $T(\tilde x)$ against those of the clean image $x$ under the same $a$ and $r$.

\subsection{Similarity-based Instruction Selection}
\label{sec:optimization}
Final-layer instruction representations are used to construct downstream generation conditions.
We therefore use the mean cosine similarity between these representations for the protected and original images to guide the selection of system prompt and editing instruction pairs during optimization.
For pair $a=(s,u)$, we compute the mean cosine similarity $c_{d,L}$ defined in Eq.~\eqref{eq:condition_loss} between the final-layer instruction representations of the transformed protected image $T(\tilde x)$ and those of the original image $x$ under the same $a,r$.
Higher similarity means that the instruction representations of the protected and original images are more closely aligned.
We compute an exponential moving average of $c_{d,L}$ for each pair.
Pairs with higher moving averages are more likely to be selected for perturbation optimization.
Let $m_{a,t}$ denote the historical score of pair $a$ before selecting a pair at step $t$.
The first observation initializes the score; each subsequent selection of $a$ updates it using the observed $c_{d,L}$:
\begin{equation}
 m_{a,t+1}=\beta m_{a,t}+(1-\beta)c_{d,L},
 \label{eq:priority}
\end{equation}
where $\beta$ is the smoothing coefficient.
Pairs not selected retain their previous scores.
Appendix~\ref{sec:optimization-settings} specifies scores for unobserved pairs.

We apply temperature-scaled softmax to the score vector $\mathbf m_t$ of all pairs to obtain their sampling probabilities:
\begin{equation}
 \boldsymbol\pi_t=\operatorname{softmax}\!\left(\mathbf m_t/\tau_s\right).
 \label{eq:entropy-sampling}
\end{equation}
Each component of $\boldsymbol\pi_t$ is the probability of selecting the corresponding pair.
The temperature $\tau_s>0$ controls the sensitivity of the selection distribution to score differences.
Lower temperatures concentrate selection on pairs with higher historical scores, while higher temperatures make the distribution more uniform.
Initially, all pairs have equal selection probabilities.
Appendix~\ref{sec:entropy-sampling-analysis} derives this distribution.
At each step, we select one $(a,r,T)$ according to $\mu_t$ and compute the loss and final-layer instruction similarity at the current perturbation $\delta_t$.
The multilayer loss and pixel regularizer update the perturbation, while $c_{d,L}$ from the same step updates the historical score and determines the next selection distribution $\boldsymbol\pi_{t+1}$.
The current probability distribution remains fixed during loss-gradient computation.
Algorithm~\ref{alg:protection} gives the complete optimization procedure with step size $\eta_t$.

\begin{algorithm}[t]
\caption{\method\ protected-image construction}
\label{alg:protection}
\small
\begin{algorithmic}[1]
\Require Image $x$, frozen VLM $F_\theta$, sets $\mathcal A,\mathcal R$,
\Statex \hspace{\algorithmicindent} layers $\Lambda$, budget $\varepsilon$, steps $N$
\Ensure Protected image $\tilde x$
\State Cache original-image hidden states for each $a,r,\ell$.
\State Initialize $\delta_0$ with small random noise.
\State Initialize $\boldsymbol\pi_0$ uniformly and set the initial scores.
\For{$t=0,\ldots,N-1$}
  \State Set the current stage parameters.
  \State Randomly select $a$ according to $\boldsymbol\pi_t$, and choose $r,T$ at random.
  \State Evaluate $\ell_{a,t}(\delta_t;r,T)$ and record $c_{d,L}$.
  \State $g_t\gets\widehat\nabla_\delta
      [\ell_{a,t}(\delta;r,T)+\lambda_t\Omega(\delta)]_{\delta=\delta_t}$.
  \State $\delta_{t+1}\gets\clip_{[-\varepsilon,\varepsilon]}
      \!\left(\delta_t-\eta_t\operatorname{sign}(g_t)\right)$.
  \State Update $m_{a,t+1}$ using $c_{d,L}$;
      $\boldsymbol\pi_{t+1}\gets\operatorname{softmax}(\mathbf m_{t+1}/\tau_s)$.
\EndFor
\State \Return $\clip_{[0,1]}(x+\delta_N)$.
\end{algorithmic}
\end{algorithm}

\section{Experiments}
\label{sec:experiments}
We evaluate cross-model protection under unseen instructions, robustness to image transformations, key optimization components, and construction cost.

\begin{table*}[!t]
\centering
\scriptsize
\setlength{\tabcolsep}{1.2pt}
\renewcommand{\arraystretch}{1.12}
\caption{\textbf{Image editing protection performance of different methods.} Both editing models use the same 200 tasks with unseen instructions, including 120 face tasks for evaluating ISM and FDFR. Clean denotes edits of the original images. Arrows indicate the directions for stronger protection. Bold and underlining mark the best and second-best protection results in each column, respectively.}
\label{tab:cross-editor-results}
\begin{tabular*}{\textwidth}{@{\extracolsep{\fill}}l*{14}{c}@{}}
\toprule
& \multicolumn{7}{c}{Step1X-Edit} & \multicolumn{7}{c}{Qwen-Image-Edit} \\
\cmidrule(lr){2-8}\cmidrule(l){9-15}
& \multicolumn{5}{c}{All tasks} & \multicolumn{2}{c}{Face tasks}
& \multicolumn{5}{c}{All tasks} & \multicolumn{2}{c}{Face tasks} \\
\cmidrule(lr){2-6}\cmidrule(lr){7-8}\cmidrule(lr){9-13}\cmidrule(l){14-15}
Method & $E\downarrow$ & $\Delta$CLIP-Dir$\uparrow$ & FID$\uparrow$ & KID$\uparrow$ & CLIP-I$\downarrow$ & ISM$\downarrow$ & FDFR (\%)$\uparrow$
& $E\downarrow$ & $\Delta$CLIP-Dir$\uparrow$ & FID$\uparrow$ & KID$\uparrow$ & CLIP-I$\downarrow$ & ISM$\downarrow$ & FDFR (\%)$\uparrow$ \\
\midrule
Clean & 8.85 & 0.000 & -- & -- & -- & 0.60 & 22.5 & 9.29 & 0.000 & -- & -- & -- & 0.62 & 18.3 \\
AdvDM & 8.15 & 0.013 & 85.75 & -0.00021 & 0.899 & 0.58 & 25.0 & 9.16 & 0.006 & 53.93 & -0.00188 & 0.944 & 0.58 & 20.0 \\
EditShield & 8.78 & 0.013 & 76.82 & -0.00050 & 0.916 & 0.58 & 25.0 & 9.29 & 0.006 & 51.20 & -0.00224 & 0.948 & 0.61 & 20.0 \\
DeContext & 6.94 & 0.021 & 90.77 & 0.00033 & 0.835 & \textbf{0.33} & \underline{31.7} & 8.49 & \underline{0.018} & 68.12 & -0.00040 & \underline{0.901} & \underline{0.51} & 25.0 \\
NullEdit & \underline{5.70} & \underline{0.061} & \underline{94.82} & \underline{0.00305} & \underline{0.831} & 0.56 & 20.0 & \underline{8.42} & 0.017 & \underline{69.58} & \underline{0.00062} & 0.912 & 0.57 & \underline{25.8} \\
\midrule
\textbf{\method{}} & \textbf{1.10} & \textbf{0.112} & \textbf{254.34} & \textbf{0.10102} & \textbf{0.523} & \underline{0.37} & \textbf{89.2} & \textbf{2.39} & \textbf{0.114} & \textbf{164.03} & \textbf{0.02604} & \textbf{0.651} & \textbf{0.47} & \textbf{29.2} \\
\bottomrule
\end{tabular*}
\end{table*}

\subsection{Experimental Setup}
\label{sec:experimental-setup}
\paragraph{Datasets and models.}
The main evaluation uses 30 images each from CelebA-HQ~\cite{celebahq} and VGGFace2~\cite{vggface2}, and 40 from GEdit-Bench~\cite{step1x}.
Each image has two test instructions held out from perturbation optimization, yielding 200 tasks.
\method{} optimizes perturbations using only a pretrained Qwen2.5-VL~\cite{qwen25vl}, keeping its parameters fixed and requiring no access to downstream generative model parameters.
Each original image has one protected image, which is directly reused on Step1X-Edit~\cite{step1x} and Qwen-Image-Edit~\cite{qwenimage}.
We use version 2509 of Qwen-Image-Edit; both models use Qwen2.5-VL.

\paragraph{Baselines.}
We select four baselines spanning image feature extraction, semantic conditioning, and diffusion generation in the editing pipeline described in Section~\ref{sec:vlm-editing}.
EditShield~\cite{editshield} and AdvDM~\cite{advdm} both optimize perturbations against the Step1X-Edit VAE.
NullEdit~\cite{nulledit} steers the protected image's VLM hidden states toward a no-edit reference, while DeContext~\cite{decontext} weakens the DiT's attention to the reference image.
We evaluate each method's protected images on the same tasks in both models, using the corresponding edits of the original images as clean references (Clean).
Appendix~\ref{sec:baseline-details} details the implementation sources, optimization settings, and computational budgets of the baselines.

\paragraph{Evaluation metrics.}
We use Qwen3.8-27B to score edited outputs from 0 to 10, assessing whether they implement the requested changes to the original image while retaining content that the instruction requires to remain.
Instruction adherence $E$ is the mean final score across test tasks; lower values indicate stronger protection.
$\Delta$CLIP-Dir measures the reduction in editing-direction agreement caused by protection.
To evaluate how protection affects edited outputs, we use Clean outputs from the same model as references, measuring overall distribution differences with FID and KID and content similarity between outputs of the same task with CLIP-I.
For face tasks, we additionally report identity similarity (ISM) and the face detection failure rate (FDFR) to evaluate identity preservation and face detection, respectively.
We evaluate protected-image fidelity with PSNR, SSIM, and LPIPS by comparing protected images with their originals.
Appendix~\ref{sec:evaluation-details} details metric computation, aggregation, and the instruction-adherence scoring protocol.

\paragraph{Implementation details.}
The candidate set for perturbation optimization contains all 60 pairings of 6 system prompts and 10 editing instructions.
We optimize each image for 3,000 steps with a perturbation budget of $\varepsilon=16/255$.
Based on the slower decrease of the instruction loss during VLM optimization, we set the image-to-instruction loss weight ratio to $4{:}6$; Section~\ref{sec:ablation-eval} provides the analysis.
During evaluation, each model uses its own system prompt, and Clean and protected results for the same task share the editing instruction, generation seed, and inference parameters.
Image-transformation experiments use a fixed set of 50 tasks, applying the same compression or resizing to the original and previously generated protected images before passing them to each model.
Appendix~\ref{sec:experimental-details} provides detailed settings for perturbation optimization and editing evaluation, and Appendix~\ref{sec:robustness-ablation-settings} gives the settings and detailed results of the robustness and ablation experiments.

\begin{table}[!t]
\centering
\scriptsize
\renewcommand{\arraystretch}{1.10}
\caption{\textbf{Protected-image fidelity.} We use each method's protected images from the main experiment to assess how perturbations affect the visual quality of protected images. Bold and underlining mark the best and second-best results in each column, respectively.}
\label{tab:published-fidelity}
\begin{tabular*}{\columnwidth}{@{\extracolsep{\fill}}lccc@{}}
\toprule
Method & PSNR (dB) $\uparrow$ & SSIM $\uparrow$ & LPIPS $\downarrow$ \\
\midrule
AdvDM & 26.16 & 0.473 & \underline{0.422} \\
EditShield & \textbf{29.44} & \textbf{0.699} & 0.443 \\
DeContext & 26.03 & 0.560 & 0.532 \\
NullEdit & 27.83 & 0.667 & 0.438 \\
\midrule
\textbf{\method{}} & \underline{28.33} & \underline{0.681} & \textbf{0.402} \\
\bottomrule
\end{tabular*}
\end{table}

\begin{table}[!t]
\centering
\scriptsize
\setlength{\tabcolsep}{1pt}
\renewcommand{\arraystretch}{1.08}
\caption{\textbf{Editing suppression after JPEG compression and resizing.} Each method reuses its protected images across transformations. Original and protected images undergo the same processing before being passed to the model. We report mean instruction adherence $E\downarrow$ on the same 50 tasks; Q denotes JPEG quality. None denotes no compression or resizing. Bold and underlining mark the best and second-best protection results in each column, respectively. Full results are provided in Table~\ref{tab:robustness-full}.}
\label{tab:processing-results}
\begin{tabular*}{\columnwidth}{@{\extracolsep{\fill}}l*{10}{c}@{}}
\toprule
& \multicolumn{5}{c}{Step1X-Edit} & \multicolumn{5}{c}{Qwen-Image-Edit} \\
\cmidrule(lr){2-6}\cmidrule(l){7-11}
Method & None & Q90 & Q80 & $0.7\times$ & $1.2\times$
       & None & Q90 & Q80 & $0.7\times$ & $1.2\times$ \\
\midrule
Clean & 9.38 & 9.40 & 9.26 & 8.98 & 9.28 & 9.38 & 9.40 & 9.38 & 9.18 & 9.40 \\
AdvDM & 8.84 & 9.24 & 9.38 & 8.66 & 8.64 & 9.08 & 9.30 & 9.30 & 8.88 & 9.32 \\
EditShield & 9.20 & 9.04 & 9.30 & 9.22 & 9.22 & 9.32 & 9.04 & 9.32 & 9.30 & 9.40 \\
DeContext & 7.66 & 8.44 & 8.48 & 8.40 & 7.68 & 8.74 & 8.72 & 8.58 & \underline{8.44} & \underline{8.34} \\
NullEdit & \underline{5.08} & \underline{7.58} & \underline{8.40} & \underline{5.90} & \underline{5.18} & \underline{8.58} & \underline{8.44} & \underline{8.50} & 8.46 & 8.50 \\
\midrule
\textbf{\method{}} & \textbf{1.30} & \textbf{2.20} & \textbf{2.10} & \textbf{2.18} & \textbf{1.38} & \textbf{2.00} & \textbf{4.34} & \textbf{6.52} & \textbf{2.28} & \textbf{1.88} \\
\bottomrule
\end{tabular*}
\end{table}

\subsection{Overall Evaluation}
\label{sec:cross-editor-results}

\method{} constructs protected images using only a pretrained VLM and evaluates them on both models using editing instructions held out from perturbation optimization, achieving the strongest editing suppression (Table~\ref{tab:cross-editor-results}).
On Step1X-Edit and Qwen-Image-Edit, $E$ is 1.10 and 2.39, respectively, 80.7\% and 71.6\% lower than the strongest baseline, NullEdit.
In Figure~\ref{fig:qualitative-overall}, several requested edits still succeed under baseline protection, whereas \method{} causes severe distortion, subject replacement, or incorrect target attributes in the outputs.
Relative to the strongest baseline for each metric, $\Delta$CLIP-Dir increases by at least 83.6\% and FID by more than 100\% across the two models.
\method{} also achieves the highest KID and lowest CLIP-I, indicating larger differences from Clean in both output distribution and image content.
AdvDM and EditShield, which optimize perturbations solely against the Step1X-Edit VAE, provide almost no protection on either model, indicating that perturbing VAE reference features alone is insufficient to suppress VLM-conditioned editing.
Among face tasks on Step1X-Edit, 89.2\% of the edited outputs from \method{} have no detected face, exceeding the 31.7\% of DeContext, the strongest baseline on this metric.
On Qwen-Image-Edit, \method{} also achieves the highest face detection failure rate and the lowest mean identity similarity among valid pairs.
In Table~\ref{tab:published-fidelity}, \method{} achieves the lowest LPIPS among all methods, while improving PSNR by 0.50--2.30~dB and SSIM by 0.014--0.208 over AdvDM, DeContext, and NullEdit.

\subsection{Protection Robustness Evaluation}
\label{sec:processing-eval}
Perturbation optimization incorporates JPEG quality factors down to 80 (Appendix~\ref{sec:perturbation-optimization}).
We evaluate robustness on the same 50 tasks by reusing the constructed protected images and applying identical compression or resizing to original and protected images (Appendix~\ref{sec:transformation-robustness-details}).
The protected images achieve the lowest $E$ in all eight model--transformation settings (Table~\ref{tab:processing-results}, Figures~\ref{fig:qualitative-jpeg} and~\ref{fig:qualitative-resize}).
Under resizing and JPEG Q90, $E$ is at least 63.1\% and 48.6\% lower than the corresponding strongest baseline, respectively.
Under stronger compression at Q80, \method{} still reduces $E$ by 75.0\% on Step1X-Edit and 23.3\% on Qwen-Image-Edit relative to NullEdit.
This advantage also extends to editing direction and output distributions: across all transformation conditions, $\Delta$CLIP-Dir and FID exceed their respective strongest baselines by at least 68.7\% and 72.0\%, respectively (Table~\ref{tab:robustness-full}).

\FloatBarrier

\subsection{Discussion}
\label{sec:ablation-eval}
\label{sec:layer-selection}
\label{sec:computational-cost}
\label{sec:efficiency-eval}
\begin{table}[!t]
\centering
\scriptsize
\setlength{\tabcolsep}{1pt}
\renewcommand{\arraystretch}{1.08}
\caption{\textbf{Protection performance of different layer combinations.} Using the configuration with only the layer-28 representation loss as the baseline, we compare protection under the same optimization budget while varying the number and positions of intermediate layers included in the loss. Bold and underlining mark the best and second-best configurations in each column, respectively.}
\label{tab:ablation-results}
\begin{tabular*}{\columnwidth}{@{\extracolsep{\fill}}l*{6}{c}@{}}
\toprule
& \multicolumn{3}{c}{Step1X-Edit} & \multicolumn{3}{c}{Qwen-Image-Edit} \\
\cmidrule(lr){2-4}\cmidrule(l){5-7}
Layers & $E\downarrow$ & $\Delta$CLIP-Dir$\uparrow$ & FID$\uparrow$
& $E\downarrow$ & $\Delta$CLIP-Dir$\uparrow$ & FID$\uparrow$ \\
\midrule
Clean & 9.26 & 0.000 & -- & 9.66 & 0.000 & -- \\
28 & 5.37 & 0.065 & 211.12 & 6.07 & 0.056 & 197.90 \\
16, 28 & \textbf{2.23} & 0.107 & 302.91 & \textbf{3.40} & 0.093 & \underline{236.24} \\
8, 16, 28 & 2.66 & \textbf{0.118} & 290.09 & 4.62 & 0.092 & 206.72 \\
8, 16, 24, 28 & \underline{2.32} & 0.112 & \underline{312.34} & 3.93 & \underline{0.102} & \textbf{243.83} \\
4, 8, 12, 28 & 2.36 & \underline{0.113} & \textbf{319.18} & 4.11 & \underline{0.102} & 227.26 \\
20, 24, 26, 28 & 4.33 & 0.079 & 249.06 & 5.49 & 0.064 & 217.89 \\
7, 14, 21, 28 & 2.52 & 0.107 & 308.64 & \underline{3.76} & \textbf{0.104} & 228.13 \\
\bottomrule
\end{tabular*}
\end{table}

\paragraph{Layer selection.}
Under the same optimization budget, all six configurations with intermediate-layer supervision improve all three metrics over final-layer supervision alone on both models (Table~\ref{tab:ablation-results}).
Even adding a single intermediate layer ($\{16,28\}$) reduces $E$ by 58.5\% on Step1X-Edit and 44.0\% on Qwen-Image-Edit, while the default $\{8,16,24,28\}$ further strengthens directional and distributional disruption ($\Delta$CLIP-Dir and FID) on both models.
Moreover, at the same layer count, $\{8,16,24,28\}$ outperforms the late-layer set $\{20,24,26,28\}$ across all metrics, confirming the benefit of spanning earlier and middle layers.
On held-out instructions, final-layer supervision substantially reduces image similarity while instruction similarity remains positive (Figure~\ref{fig:layer-ablation}; Appendix~\ref{sec:ablation-details}).
Adding intermediate-layer supervision lowers instruction similarity further.
The default configuration produces larger directional changes in final-layer instruction representations than the later-layer configuration, consistent with stronger editing suppression (Figure~\ref{fig:qualitative-layers}).

\begin{figure}[!t]
\centering
\includegraphics[width=0.92\columnwidth]{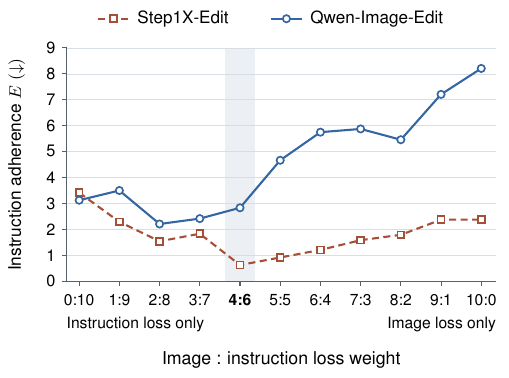}
\caption{\textbf{Loss weight ablation.} Mean instruction adherence $E$ across 11 weight ratios. Shading marks the default $4{:}6$.}
\label{fig:weight-ablation}
\end{figure}

\paragraph{Loss weights.}
When we optimize the perturbation with both losses, instruction similarity decreases more slowly than image similarity.
We therefore give the instruction term more weight, using an image-to-instruction ratio of $4{:}6$ (Appendix~\ref{sec:ablation-details}).
Figure~\ref{fig:weight-ablation} shows that the default $4{:}6$ yields lower $E$ than either loss alone on both editors.
Its mean $E$ of 1.73 across the two models is the lowest among all ratios, 67.3\% and 47.1\% below the image-only and instruction-only settings, respectively.
$E$ is lowest at $4{:}6$ for Step1X-Edit and $2{:}8$ for Qwen-Image-Edit.
Figure~\ref{fig:qualitative-weights} compares edited outputs across loss weights.

\paragraph{Computational cost.}
\begin{table}[!tb]
\centering
\scriptsize
\setlength{\tabcolsep}{6pt}
\renewcommand{\arraystretch}{1.18}
\caption{\textbf{Protected-image construction cost.} Average time to construct one protected image and peak GPU memory usage for each method. All measurements are taken on an NVIDIA RTX PRO 6000 Blackwell GPU.}
\label{tab:construction-cost}
\begin{tabular}{@{}lcc@{}}
\toprule
Method & Time (s/image) $\downarrow$ & Peak memory (GiB) $\downarrow$ \\
\midrule
AdvDM & \makebox[3.2em][r]{51.9} & \makebox[3.2em][r]{15.59} \\
EditShield & \makebox[3.2em][r]{100.0} & \makebox[3.2em][r]{13.16} \\
DeContext & \makebox[3.2em][r]{1194.9} & \makebox[3.2em][r]{56.71} \\
NullEdit & \makebox[3.2em][r]{1793.8} & \makebox[3.2em][r]{21.03} \\
\midrule
\textbf{\method{}} & \makebox[3.2em][r]{584.7} & \makebox[3.2em][r]{20.59} \\
\bottomrule
\end{tabular}
\end{table}

Relative to NullEdit, which also constructs protection by perturbing VLM representations, \method{} reduces protected-image construction time by 67.4\% at similar peak memory and achieves stronger editing suppression on both models (Tables~\ref{tab:cross-editor-results} and~\ref{tab:construction-cost}).
\method{} also constructs protected images faster than DeContext while reducing peak memory by 63.7\%.
Although AdvDM and EditShield run faster by optimizing only against the VAE, they provide almost no protection on either editing model.

\section{Conclusion}
\label{sec:conclusion}
\method{} constructs transferable image protection by disrupting both image and instruction representations of a frozen pretrained VLM.
Multilayer supervision and similarity-based instruction selection guide optimization without accessing downstream generative model parameters.
The same protected images reduce instruction adherence by at least 71.6\% relative to the strongest baseline on each editor, achieve lower LPIPS than all baselines, and retain protection after JPEG compression and resizing.
These results establish the shared VLM encoding as an effective basis for protection across editors.

\FloatBarrier
\section*{LLM Usage Statement}
We used OpenAI Codex and DeepSeek to support code development and to edit the wording of manuscript text drafted by the authors.

{
\small
\bibliographystyle{ieeenat_fullname}
\bibliography{references}
}

\clearpage
\appendix
\section{Analysis of Adaptive Instruction Selection}
\label{sec:adaptive-analysis}

\subsection{Joint Sampling Distribution}
\label{sec:effective-distribution}
Let $\nu_t(r,T)$ denote the joint distribution of the visual-token grid and image transformation at step $t$.
Given the historical scores, we select pair $a=(s,u)$ according to $\boldsymbol\pi_t$, independently of $(r,T)$.
The joint distribution in Section~\ref{sec:method-formulation} can therefore be written as
\begin{equation}
 \mu_t(a,r,T)=\pi_{a,t}\,\nu_t(r,T).
 \label{eq:joint-condition-distribution}
\end{equation}
At each step, we estimate the expected representation loss by evaluating $\ell_{a,t}(\delta;r,T)$ at one sampled triple $(a,r,T)$~\cite{ji2026anglerocl}.
We hold $\mu_t$ fixed while differentiating the sampled loss with respect to $\delta$.
Backpropagation through JPEG and pixel quantization uses the surrogate operators in Appendix~\ref{sec:perturbation-optimization}, giving the update gradient $\widehat\nabla_\delta$ in Algorithm~\ref{alg:protection}.

\subsection{Entropy-Regularized Instruction Selection}
\label{sec:entropy-sampling-analysis}
The softmax distribution in Eq.~\eqref{eq:entropy-sampling} balances prioritizing pairs with higher historical instruction similarity against maintaining sampling diversity.
Given the score vector $\mathbf m_t$ and temperature $\tau_s>0$, this distribution solves the following entropy-regularized objective over the probability simplex $\Delta(\mathcal A)$:
\begin{equation}
 \max_{\boldsymbol\pi\in\Delta(\mathcal A)}
 \left\{\sum_{a\in\mathcal A}\pi_a m_{a,t}
       -\tau_s\sum_{a\in\mathcal A}\pi_a\log\pi_a\right\}.
\end{equation}
The objective is strictly concave in $\boldsymbol\pi$, so the optimal distribution is unique.
Let $\gamma$ be the Lagrange multiplier for the normalization constraint $\sum_a\pi_a=1$.
The first-order condition is
\begin{equation}
 m_{a,t}-\tau_s(\log\pi_a+1)-\gamma=0.
\end{equation}
Applying the normalization constraint gives
\begin{equation}
 \pi_{a,t}
 =\frac{\exp(m_{a,t}/\tau_s)}
 {\sum_{b\in\mathcal A}\exp(m_{b,t}/\tau_s)}.
\end{equation}
For any $a,b\in\mathcal A$, the ratio of their selection probabilities is
$\pi_{a,t}/\pi_{b,t}=\exp((m_{a,t}-m_{b,t})/\tau_s)$.
Lowering the temperature concentrates selection on pairs with higher historical instruction similarity, whereas raising it moves the distribution toward uniform.

\section{Experimental Details}
\label{sec:experimental-details}
\subsection{\method{} Perturbation Optimization}
\label{sec:perturbation-optimization}
\label{sec:optimization-settings}
We instantiate $F_\theta$ with frozen Qwen2.5-VL-7B~\cite{qwen25vl} and use $\varepsilon=16/255$ and $\Lambda=\{8,16,24,L\}$, where the final layer is $L=28$.
We initialize the perturbation with small random noise.
The loss weights for image and instruction representations are $(w_v,w_d)=(0.4,0.6)$.
Each step selects one $(a,r,T)$, updates the perturbation using the gradient sign, and projects it onto the perturbation budget.
We encode the original and protected images with the same system prompt, editing instruction, and visual-token grid, and compare their hidden states at corresponding token positions.
The image term uses image tokens, and the instruction term uses valid text positions after the image-end marker.
Similarities are aggregated with the layer weights to form the image and instruction representation losses.
Optimization uses 3,000 steps divided into the four stages in Table~\ref{tab:schedule}.
The three intermediate layers equally share the weight remaining after the final-layer weight is assigned.
The initial step sizes for the four stages are $1/255$, $0.6/255$, $0.4/255$, and $0.2/255$.
Within each stage, the step size is multiplied by 0.7 every 80 steps and reset at the stage transition.

All combinations of six system prompts and ten user editing instructions form 60 pairs $a=(s,u)$ in the training set $\mathcal A$.
Tables~\ref{tab:optimization-system-prompts} and~\ref{tab:optimization-user-instructions} list the full texts.
The system prompts request the specified edit while preserving unrelated content, and the editing instructions cover brightness and color adjustments, background changes, style transfer, and scene lighting changes.
Initially, all pairs have the same chance of being selected.
The first observation of final-layer instruction similarity $c_{d,L}$ initializes a pair's score. We update that score with an exponential moving average ($\beta=0.9$, Eq.~\eqref{eq:priority}) and sample pairs by softmax ($\tau_s=0.2$, Eq.~\eqref{eq:entropy-sampling}).
Unselected pairs retain their scores, and unobserved pairs use the mean observed score, or zero when no scores have been observed.
The score at step $t$ is determined by representation similarity at $\delta_t$ and used for selection at the next step.

\begin{table}[ht]
\centering
\small
\renewcommand{\thetable}{B.1}
\caption{\textbf{Perturbation optimization schedule.} Processing probabilities are listed in the order of identity, resizing, and JPEG compression. $\alpha_{L,t}$ is the final-layer loss weight, and $\lambda_t$ is the pixel regularization weight.}
\label{tab:schedule}
\renewcommand{\arraystretch}{1.08}
\begin{tabular*}{\columnwidth}{@{\extracolsep{\fill}}ccccc@{}}
\toprule
Stage & Steps & \shortstack{Processing\\probabilities} & $\alpha_{L,t}$ & $\lambda_t$ \\
\midrule
1 & 800 & $(1,0,0)$ & $1/4$ & $0.05$ \\
2 & 800 & $(0.2,0.8,0)$ & $1/4$ & $0$ \\
3 & 800 & $(0.2,0.4,0.4)$ & $1/4$ & $0$ \\
4 & 600 & $(0.2,0.4,0.4)$ & $1/2$ & $0$ \\
\bottomrule
\end{tabular*}
\end{table}

At each optimization step, we choose a target area of $384^2$ or $512^2$ pixels with equal probability and compute the corresponding dimensions from the original aspect ratio.
We generate candidate sizes by rounding these dimensions to the nearest multiple of 16 or 32, or down to a multiple of 16, then aligning each candidate to the $28\times28$-pixel grid.
We sample uniformly from the distinct grid sizes. To compute the reference representation, we resize the original image directly to the sampled size using bicubic interpolation.
For the resizing transformation, we express the intermediate scale as the side length $S$ of an equivalent square.
We draw $S$ from $384,416,\ldots,992$ for the $384^2$-pixel target and from $512,544,\ldots,992$ for the $512^2$-pixel target according to
\[
 w(S)=1+2\frac{S_{\max}-S}{S_{\max}-S_{\min}},
 \qquad P(S)=\frac{w(S)}{\sum_{S'}w(S')}.
\]
The smallest scale receives three times the weight of the largest.
We choose an alignment rule for the intermediate size (no alignment, nearest-16, nearest-32, or floor-16) and an interpolation method (Lanczos, bicubic, bilinear, or box). Both choices are uniform.
We then resize the intermediate image to the selected input size with bicubic interpolation.

When applying JPEG during perturbation optimization, we sample its quality factor uniformly from $\{80,85,90,95\}$.
Lower quality factors can be included when stronger compression is expected.
The JPEG transformation uses Pillow encoding and decoding in the forward pass and BPDA~\cite{bpda} to estimate gradients in the backward pass.
The surrogate gradient for JPEG quantization combines the straight-through estimator (STE, weight 0.25) with the gradient of a cubic surrogate function (weight 0.75).
For a value $z$ to be rounded, the cubic surrogate is $\operatorname{round}(z)+(z-\operatorname{round}(z))^3$, treating $\operatorname{round}(z)$ as constant during differentiation.
Pixel quantization in resizing uses the STE.
Algorithm~\ref{alg:protection} denotes these surrogate differentiation operations by $\widehat\nabla$.
After optimization, we save the protected image at its original resolution as an 8-bit RGB PNG for subsequent editing evaluation.

\begin{table*}[t]
\centering
\small
\setlength{\tabcolsep}{4pt}
\renewcommand{\arraystretch}{1.15}
\renewcommand{\thetable}{B.2}
\caption{\textbf{System prompts used for perturbation optimization.}}
\label{tab:optimization-system-prompts}
\begin{tabular}{@{}lp{0.93\textwidth}@{}}
\toprule
ID & System prompt \\
\midrule
S1 & You are a visual editing assistant. Use the provided image and the user's request to produce an edited result. Follow the requested change precisely and keep unrelated content stable. \\
S2 & Examine the visual scene, identify the relevant objects, attributes, and spatial relationships, and then carry out the user's requested modification. Avoid introducing changes that are not required by the instruction. \\
S3 & Treat the user's instruction as the primary editing objective. Modify the supplied image accordingly while preserving visual coherence, realism, and the identity of regions that are not involved in the requested edit. \\
S4 & You are a multimodal assistant capable of understanding images and executing visual transformation requests. Generate a result that reflects the user's intent and remains consistent with the source image. \\
S5 & Edit the input image according to the user's request. Preserve everything else whenever possible. \\
S6 & Perform the requested image transformation using the source image as the reference. Maintain the original composition, subject identity, lighting, and overall quality unless the user explicitly asks for those elements to change. \\
\bottomrule
\end{tabular}
\end{table*}

\begin{table*}[t]
\centering
\small
\setlength{\tabcolsep}{4pt}
\renewcommand{\arraystretch}{1.15}
\renewcommand{\thetable}{B.3}
\caption{\textbf{User editing instructions used for perturbation optimization.}}
\label{tab:optimization-user-instructions}
\begin{tabular}{@{}lp{0.93\textwidth}@{}}
\toprule
ID & User editing instruction \\
\midrule
U1 & Increase the overall brightness while preserving all important visual details. \\
U2 & Apply a warm cinematic color tone without changing the main subjects or composition. \\
U3 & Make the background softly blurred while keeping the foreground subject sharp and natural. \\
U4 & Convert the image into a detailed watercolor painting with soft edges and subtle paper texture. \\
U5 & Adjust the scene to look as if it were photographed during golden hour, with gentle warm sunlight. \\
U6 & Replace the existing background with a clean studio setting while maintaining the original subject, pose, proportions, and shadows. \\
U7 & Remove small distracting objects from the scene and reconstruct the missing areas so that the result looks visually consistent and realistic. \\
U8 & Add soft natural depth of field, improve facial and object clarity, and keep the original colors, layout, identity, and overall atmosphere unchanged. \\
U9 & Transform the scene into a realistic nighttime version with believable artificial lighting, deeper shadows, visible highlights, and consistent illumination across every object and person. \\
U10 & Restyle the entire image as a high-quality cinematic photograph. Improve the lighting, color balance, texture detail, and visual depth while preserving the identity of the subject and the original spatial relationships. \\
\bottomrule
\end{tabular}
\end{table*}

\subsection{Baseline Implementations}
\label{sec:baseline-details}
We configure the baselines based on their official implementations or descriptions in the original papers, adapting them to the corresponding components of Step1X-Edit v1p1.
All methods use an $\ell_\infty$ perturbation budget of $\varepsilon=16/255$ in RGB pixel space normalized to $[0,1]$.
We adapt AdvDM~\cite{advdm} to the Step1X-Edit VAE by replacing the diffusion objective with the norm of the VAE latent representation, which we maximize using 200 PGD steps with an initial step size of $1/255$ and linear decay.
After optimization, we scale the perturbation by a factor of two and project it back onto the perturbation budget.
For EditShield~\cite{editshield}, we maximize the mean of the latent-representation MSE terms for the noisy and conditioning branches, minus 0.1 times the pixel MSE.
We use Adam for 300 steps with a fixed learning rate of $10^{-3}$. Optimization starts from the original image, with projection onto the perturbation budget after each update.

DeContext~\cite{decontext} uses the native Step1X-Edit v1p1 editing forward pass at high-noise timesteps to suppress attention from generated tokens to reference-image tokens.
The objective operates on single blocks 0--25, and optimization uses 60 editing instructions. We run 200 steps, using a step size of $0.005$ for the first 100 steps and $0.0035$ for the remaining 100.
NullEdit~\cite{nulledit} uses the VLM text encoder and its preprocessing, steering the protected-image representation under an editing instruction away from that of the original image under the same instruction and toward that of the original image under a no-edit instruction.
We run 1,000 steps with an initial step size of $1/255$ and cosine decay. The momentum coefficient is 0.9, and the no-edit term has weight 1.
Each iteration averages gradients over eight editing instructions.
AdvDM, EditShield, and DeContext optimize pixel perturbations at the original image resolution, with native model preprocessing resizing inputs to an area of approximately $1024^2$ pixels.
On the same hardware, DeContext and NullEdit take approximately 2.04 and 3.07 times as long as \method{}, respectively, to construct a protected image (Table~\ref{tab:construction-cost}).

\subsection{Evaluation Protocol and Metrics}
\label{sec:evaluation-details}
The main evaluation includes 30 images each from CelebA-HQ and VGGFace2, and 40 from GEdit-Bench.
Each image has two test instructions excluded from \method{} perturbation optimization, yielding 200 editing tasks, of which 120 are face tasks.
Each method constructs one protected image per original and reuses it in Step1X-Edit and Qwen-Image-Edit.
The two models use their own system prompts and inference settings.
Within each model, the original and protected images for a task share the editing instruction, generation seed, and inference parameters.
We denote the edited output of the original image as Clean.
Qwen3.8-27B scores each edit against the original image and instruction on an integer scale from 0 to 10. An edit earns 10 if it completes the request on the intended target and preserves the required content, and 0 if it makes no meaningful requested change or yields an unusable image. For initial scores above 3, the model checks for damage to content that should remain. Even a completed edit earns at most 3 if a required central element is lost or replaced, and at most 5 if required content remains recognizable after a major unrequested redraw. The initial score stands if the review finds the required content intact or explains why the visual evidence is inconclusive. If the review lacks concrete evidence or a valid judgment, the model scores the edit again using the original instruction. In the main evaluation, $E$ is the mean final score over 200 tasks for each method and editor.

CLIP direction agreement measures the alignment between image changes and corresponding description changes in CLIP feature space.
For $\Delta$CLIP-Dir, we subtract the protected output's direction agreement from that of Clean for each task and average over the 200 tasks.
A larger $\Delta$CLIP-Dir indicates a larger drop in directional agreement relative to Clean.
FID and kernel inception distance (KID) compare the feature distribution of a method's 200 edited outputs with that of the 200 Clean outputs from the same model.
FID computes the Fr\'echet distance from the means and covariances of Inception features, while KID computes the squared maximum mean discrepancy between these feature distributions.
CLIP image similarity (CLIP-I) computes the cosine similarity between the CLIP image features of each protected output and its corresponding Clean output, then averages across tasks.
Higher FID and KID and lower CLIP-I indicate greater deviation from Clean in the overall output distribution and in image content within the same task, respectively.

On the 120 face tasks, ISM computes the cosine similarity between ArcFace features of the original image and edited result, averaged only over pairs with successfully extracted features for both images.
FDFR is the proportion of the 120 face tasks whose edited outputs have no detected face.
Lower ISM and higher FDFR indicate weaker identity preservation and more frequent face-detection failures, respectively.
Protected-image fidelity directly compares original and protected images, averaging PSNR, SSIM, and LPIPS over 100 image pairs per method.
Peak signal-to-noise ratio (PSNR) measures fidelity based on pixel mean squared error, structural similarity (SSIM) compares local luminance, contrast, and structure, and LPIPS measures perceptual differences through deep feature distances.
Higher PSNR and SSIM and lower LPIPS indicate that the protected image is closer to the original.

\FloatBarrier
\section{Robustness and Ablation Results}
\label{sec:robustness-ablation-settings}

\subsection{Robustness to Image Transformations}
\label{sec:transformation-robustness-details}
\begin{table*}[!t]
\centering
\small
\setlength{\tabcolsep}{3pt}
\renewcommand{\arraystretch}{1.13}
\renewcommand{\thetable}{C.1}
\caption{\textbf{Editing direction and output distribution under image transformations.} We report $\Delta$CLIP-Dir and FID on the 50 tasks in Table~\ref{tab:processing-results}, using edits of identically processed original images (Clean) as references. None denotes no additional processing, and Q denotes JPEG quality. Bold and underlining indicate the best and second-best results, respectively.}
\label{tab:robustness-full}
\begin{tabular*}{\textwidth}{@{\extracolsep{\fill}}l*{10}{c}@{}}
\toprule
& \multicolumn{5}{c}{Step1X-Edit} & \multicolumn{5}{c}{Qwen-Image-Edit} \\
\cmidrule(lr){2-6}\cmidrule(l){7-11}
Method & None & Q90 & Q80 & $0.7\times$ & $1.2\times$
       & None & Q90 & Q80 & $0.7\times$ & $1.2\times$ \\
\midrule
\multicolumn{11}{@{}l}{\textbf{$\Delta$CLIP-Dir $\uparrow$}} \\[1pt]
AdvDM & 0.009 & 0.011 & 0.002 & 0.023 & 0.016 & 0.010 & 0.011 & 0.011 & 0.010 & 0.012 \\
EditShield & 0.014 & 0.014 & 0.004 & 0.023 & 0.021 & 0.001 & 0.007 & 0.005 & 0.006 & 0.005 \\
DeContext & 0.030 & 0.025 & 0.014 & 0.022 & 0.029 & \underline{0.025} & \underline{0.024} & \underline{0.021} & \underline{0.027} & \underline{0.027} \\
NullEdit & \underline{0.088} & \underline{0.038} & \underline{0.028} & \underline{0.073} & \underline{0.083} & 0.016 & 0.022 & \underline{0.021} & 0.019 & 0.023 \\
\midrule
\textbf{\method{}} & \textbf{0.138} & \textbf{0.128} & \textbf{0.128} & \textbf{0.134} & \textbf{0.140} & \textbf{0.131} & \textbf{0.117} & \textbf{0.073} & \textbf{0.139} & \textbf{0.134} \\
\midrule
\multicolumn{11}{@{}l}{\textbf{FID $\uparrow$}} \\[1pt]
AdvDM & 110.45 & 76.64 & 74.78 & 85.38 & 95.63 & 69.37 & 62.72 & 66.98 & 62.37 & 65.73 \\
EditShield & 106.80 & 84.81 & 80.37 & 87.57 & 86.31 & 65.89 & 64.58 & 61.98 & 54.40 & 62.76 \\
DeContext & 137.80 & \underline{109.46} & \underline{102.13} & 103.59 & 121.08 & \underline{87.86} & \underline{90.79} & \underline{88.58} & 76.33 & \underline{87.63} \\
NullEdit & \underline{148.70} & 109.24 & 100.52 & \underline{144.17} & \underline{140.64} & 85.31 & 86.30 & 83.43 & \underline{87.28} & 86.01 \\
\midrule
\textbf{\method{}} & \textbf{319.11} & \textbf{279.03} & \textbf{269.31} & \textbf{300.18} & \textbf{314.33} & \textbf{212.77} & \textbf{196.97} & \textbf{152.40} & \textbf{236.62} & \textbf{218.27} \\
\bottomrule
\end{tabular*}
\end{table*}

We evaluate robustness on a fixed set of 50 tasks under JPEG compression at quality factors 90 and 80, aspect-ratio-preserving resizing by $0.7\times$ and $1.2\times$, and no additional processing.
For each method, we reuse the previously constructed protected images and apply the same transformation to the original and protected images before editing.
The evaluator uses the transformed original image and editing instruction as references, and $E$ is the mean final score over the 50 tasks for each model and condition.
$\Delta$CLIP-Dir compares Clean and protected outputs within each task and condition, while FID compares the distributions of the two sets of 50 edited outputs.
Table~\ref{tab:processing-results} reports $E$, and Table~\ref{tab:robustness-full} provides the editing-direction and output-distribution results.
Figures~\ref{fig:qualitative-jpeg} and~\ref{fig:qualitative-resize} show matched outputs for the compression and resizing conditions.

\subsection{Ablation Studies}
\label{sec:ablation-details}

\begin{figure}[!tb]
\centering
\includegraphics[width=\columnwidth]{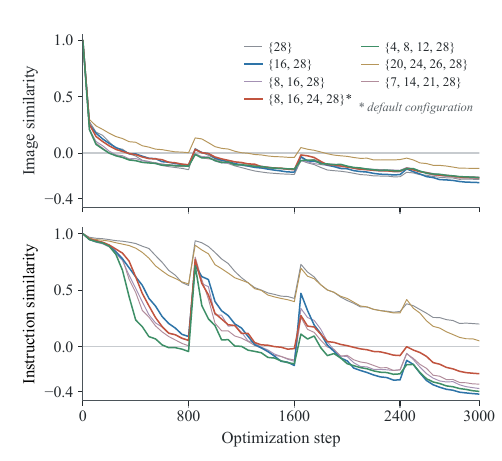}
\renewcommand{\thefigure}{C.1}
\caption{\textbf{Final-layer representation similarity on held-out instructions.} The top panel tracks image-token similarity and the bottom panel tracks instruction-token similarity. Both plot the mean cosine similarity between layer-28 hidden states at matching token positions for the original and protected images. Each curve represents a supervised-layer configuration, and the asterisk marks the default.}
\label{fig:layer-ablation}
\end{figure}

\paragraph{Layer Selection.}
We conduct the layer-selection ablation on a 50-image subset drawn from the three datasets, with two held-out editing instructions per image.
We compare seven layer configurations: $\{28\}$, $\{16,28\}$, $\{8,16,28\}$, $\{8,16,24,28\}$, $\{4,8,12,28\}$, $\{20,24,26,28\}$, and $\{7,14,21,28\}$.
Each configuration uses the small random-noise initialization in Appendix~\ref{sec:perturbation-optimization}, 3,000 optimization steps, and an image-to-instruction loss weight ratio of $4{:}6$.
All layer configurations use the same shuffled-cycle selection rule: we traverse all 60 system-prompt and editing-instruction pairs once per pass, reshuffling at the start of each pass.
The final-layer-only configuration always assigns weight 1 to that layer.
For all other configurations, the final layer receives weight $1/4$ in the first three stages, with the intermediate layers equally sharing the remaining $3/4$; in the last stage, the final layer receives $1/2$ and the intermediate layers equally share the other $1/2$.
We evaluate each of the 100 image--instruction tasks under five conditions covering unprocessed inputs, JPEG quality 90 or 70, and resizing by $0.7\times$ or $1.2\times$.
For each editor and layer configuration, $E$ is the mean final score over all 500 outputs. $\Delta$CLIP-Dir uses the Clean output for the same task and condition as its reference.
FID is computed separately on the 100 outputs in each condition and then averaged over the five conditions.
To track representation changes during optimization (Figure~\ref{fig:layer-ablation}), we fix two held-out instructions and two visual grids per image. Every 50 steps we measure layer-28 similarities without additional image transformations, averaging first over the four instruction--grid combinations and then over the 50 images.
Figure~\ref{fig:qualitative-layers} shows edited outputs for four of the layer configurations.

\paragraph{Loss Weights.}
As shown in Figure~\ref{fig:layer-ablation}, when jointly optimizing the image and instruction representation losses on the VLM, instruction similarity decreases more slowly than image similarity, motivating an image-to-instruction ratio of $4{:}6$.
The loss-weight ablation uses a 50-image subset drawn from the three datasets and compares 11 image-to-instruction ratios, $(k{:}10-k)$ for $k=0,\ldots,10$.
The endpoints $0{:}10$ and $10{:}0$ use only the instruction and image representation losses, respectively.
Across the 11 ratios, we keep all other settings fixed, including the default four-layer configuration, 3,000 optimization steps, and similarity-based instruction selection in Section~\ref{sec:optimization}.
After constructing the protected images, we evaluate one test instruction per image under no transformation, JPEG quality 90, and resizing by factors of $0.9$ and $1.1$, yielding 200 edited results per model and ratio.
Figure~\ref{fig:weight-ablation} in Section~\ref{sec:ablation-eval} plots mean $E$ for each editor. The cross-editor comparison averages these two scores with equal weight.
Figure~\ref{fig:qualitative-weights} shows edited outputs for four image-to-instruction weight ratios, including the default $4{:}6$.

\clearpage
\setcounter{figure}{0}
\renewcommand{\thefigure}{\thesection.\arabic{figure}}
\renewcommand{\theHfigure}{appendix.\thesection.\arabic{figure}}

\onecolumn
\captionsetup{hypcap=false}
\noindent\begin{minipage}{\textwidth}
  \section{Qualitative Results}
  \label{sec:qualitative-results}

  \paragraph{Overall Protection.}
  Figure~\ref{fig:qualitative-overall} complements Table~\ref{tab:cross-editor-results} with one held-out editing task each from CelebA-HQ, GEdit-Bench, and VGGFace2.
  Baselines still permit the requested edits; in Figure~\ref{fig:qualitative-overall}(b), for example, both editors turn the clean bag red, whereas with \method{} the bag is unrecognizable in Step1X-Edit and turns green in Qwen-Image-Edit.

  \medskip
  {\centering
  \includegraphics[width=\linewidth,height=0.82\textheight,keepaspectratio]{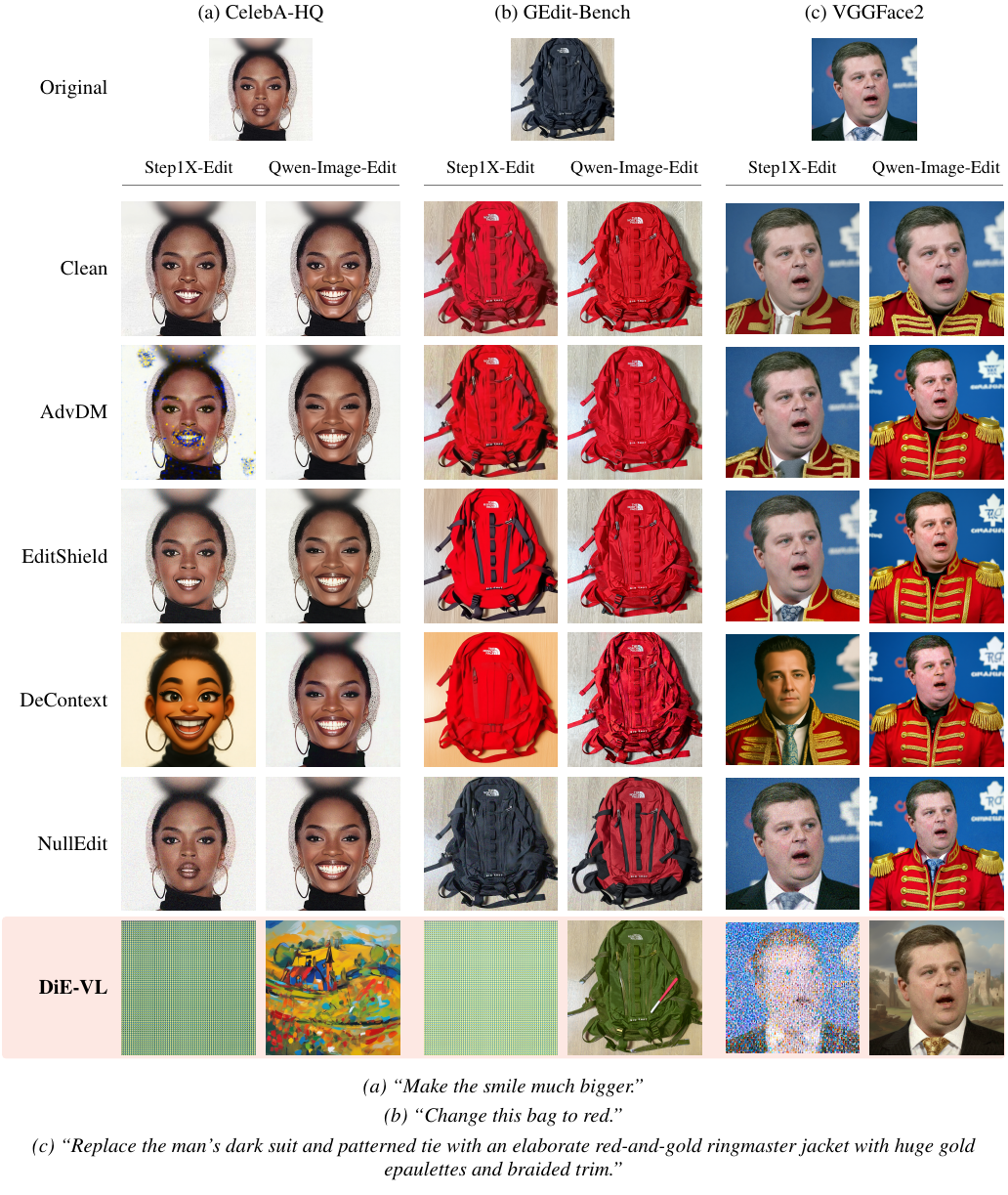}\par}
  \captionof{figure}{\textbf{Qualitative comparison of image-editing protection.} Each column group shows one task, with Step1X-Edit and Qwen-Image-Edit outputs side by side. Original shows the unprotected input image. Clean shows each editor's output for that input. Each protection method uses the same protected image across both editors.}
  \label{fig:qualitative-overall}
\end{minipage}
\clearpage

\paragraph{Robustness to Image Transformations.}
Figure~\ref{fig:qualitative-jpeg} compares JPEG settings for the task of adding a gold mask to a portrait. Figure~\ref{fig:qualitative-resize} compares resize factors for a second portrait, where the instruction asks for a crown.

\medskip
\noindent\begin{minipage}{\textwidth}
{\centering
\includegraphics[width=0.83\linewidth,height=0.82\textheight,keepaspectratio]{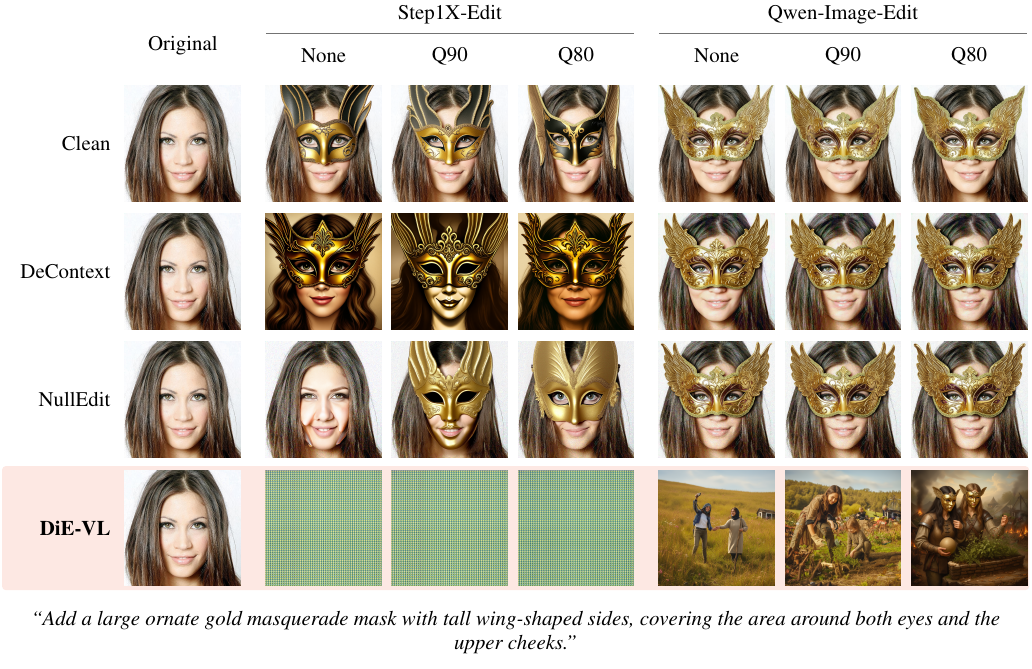}\par}
\captionof{figure}{\textbf{Editing results under JPEG compression.} For each editor, we compare outputs without additional processing (None) and after JPEG compression at quality factors 90 and 80. Original shows the unprocessed input, and Clean shows the edited original under each condition. The same processing is applied to the original and protected images before editing.}
\label{fig:qualitative-jpeg}
\end{minipage}

\medskip
\noindent\begin{minipage}{\textwidth}
{\centering
\includegraphics[width=0.83\linewidth,height=0.82\textheight,keepaspectratio]{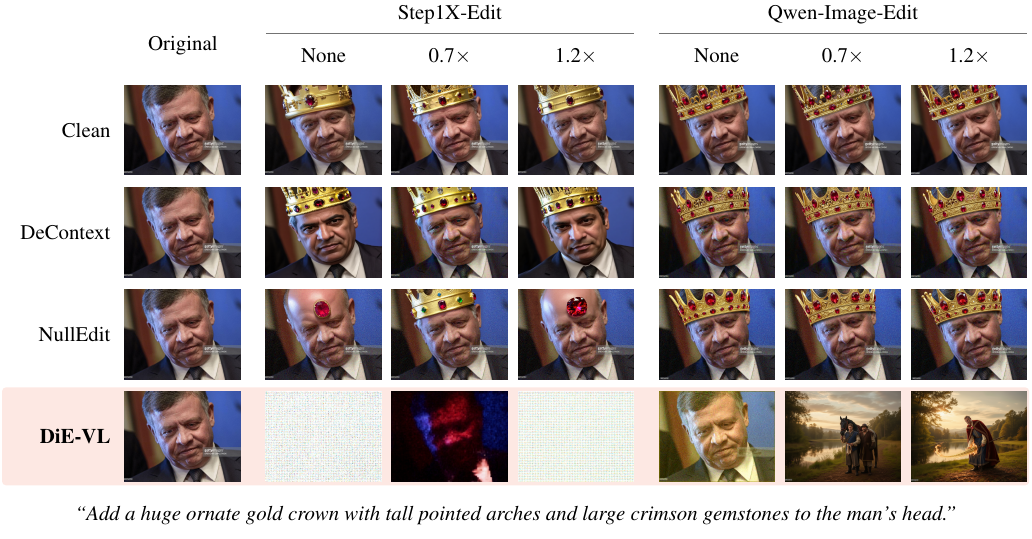}\par}
\captionof{figure}{\textbf{Editing results under resizing.} For each editor, we compare outputs without additional processing (None) and after resizing by $0.7\times$ and $1.2\times$. Original shows the unprocessed input, and Clean shows the edited original under each condition. The original and protected images are resized identically with their aspect ratios preserved.}
\label{fig:qualitative-resize}
\end{minipage}

\clearpage
\paragraph{Ablation Studies.}
Figure~\ref{fig:qualitative-layers} varies the supervised layers. Figure~\ref{fig:qualitative-weights} varies the image-to-instruction loss weight ratio.
Each figure shows two editing tasks without additional image transformations.

\medskip
\noindent\begin{minipage}{\textwidth}
{\centering
\includegraphics[width=\linewidth,height=0.82\textheight,keepaspectratio]{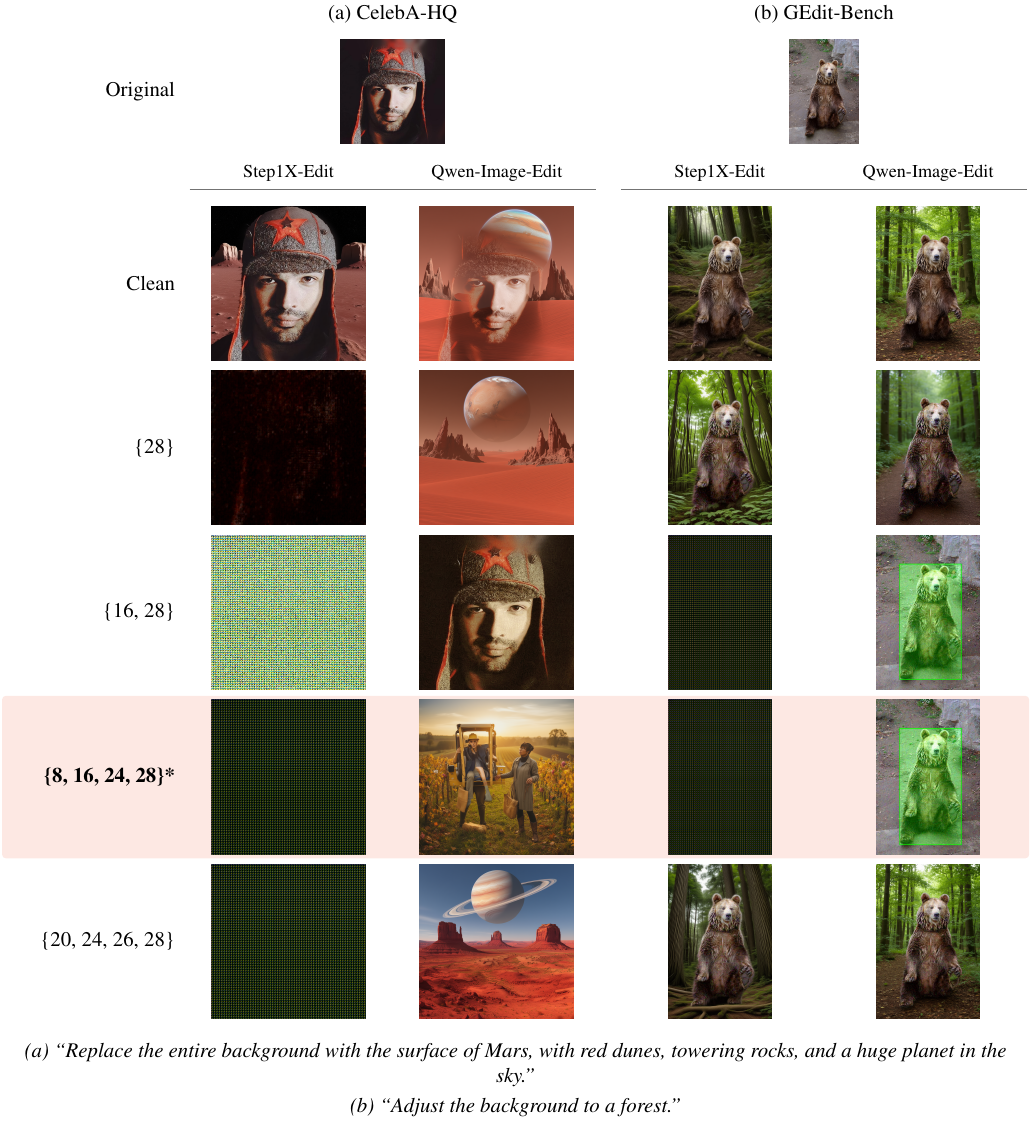}\par}
\captionof{figure}{\textbf{Editing results under different layer configurations.} Each column group shows one task, with Step1X-Edit and Qwen-Image-Edit outputs side by side. Original shows the unprotected input image. Clean shows each editor's output for that input. The remaining rows list the supervised layers. Shading and an asterisk mark the default configuration $\{8,16,24,28\}$.}
\label{fig:qualitative-layers}
\end{minipage}

\clearpage
\noindent\begin{minipage}{\textwidth}
{\centering
\includegraphics[width=\linewidth,height=0.82\textheight,keepaspectratio]{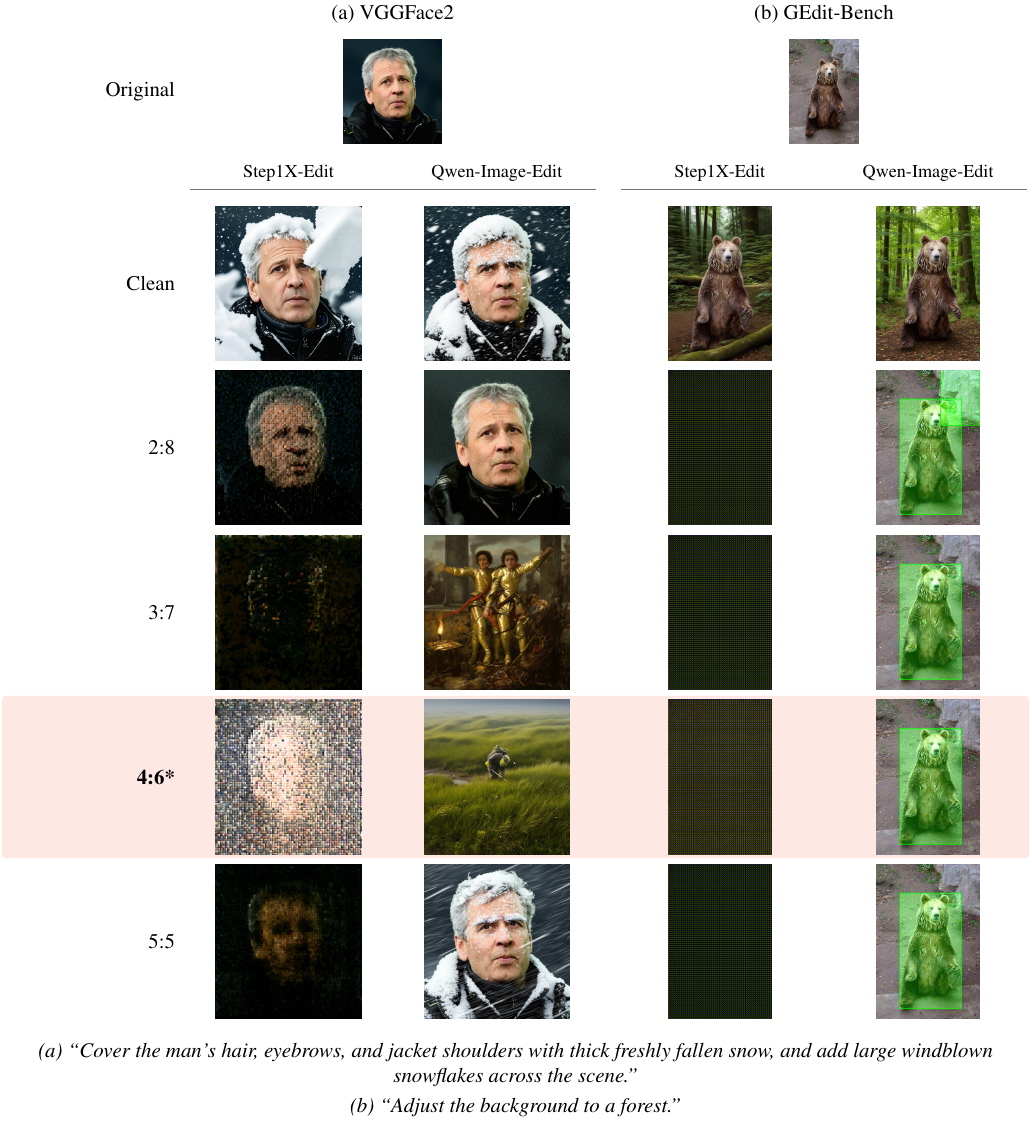}\par}
\captionof{figure}{\textbf{Editing results under different loss weights.} Each column group shows one task, with Step1X-Edit and Qwen-Image-Edit outputs side by side. Original shows the unprotected input image. Clean shows each editor's output for that input. The remaining rows give the image-to-instruction loss weight ratios. Shading and an asterisk mark the default ratio of $4{:}6$.}
\label{fig:qualitative-weights}
\end{minipage}

\end{document}